\documentclass{article}
\usepackage{ijcai25}

\usepackage{times}
\usepackage{soul}
\usepackage{url}
\usepackage[hidelinks]{hyperref}
\usepackage[utf8]{inputenc}
\usepackage[small]{caption}
\usepackage{graphicx}
\usepackage{amsmath}
\usepackage{amsthm}
\usepackage{booktabs}
\usepackage{algorithm}
\usepackage{algorithmic}
\usepackage[switch]{lineno}

\newtheorem{example}{Example}

\author{
Hongjia Liu
\and
Jinlong Li$^1$\\
\affiliations
$^1$University of Science and Technology of China\\
\emails
inalian@mail.ustc.edu.cn,
jlli@ustc.edu.cn
}

\title{A Training-free LLM Framework with Interaction between Contextually Related Subtasks in Solving Complex Tasks}

\usepackage{array}
\usepackage{pgfplots}
\pgfplotsset{compat=1.18}

\begin{document}

\maketitle

\begin{abstract}

Large language models (LLMs) have shown remarkable capabilities in solving complex tasks. Recent work has explored decomposing such tasks into subtasks with independent contexts. However, some contextually related subtasks may encounter information loss during execution, leading to redundant operations or execution failures. To address this issue, we propose a training-free framework with an interaction mechanism, which enables a subtask to query specific information or trigger certain actions in completed subtasks by sending requests. To implement interaction, we introduce a subtask trajectory memory to enable resumption of completed subtasks upon receiving interaction requests. Additionally, we propose a new action during execution, which generates a concise and precise description of execution process and outcomes of a subtask, to assist subsequent subtasks in determining interaction targets and requests. We evaluate our framework on interactive decision-making task WebShop and multi-hop question answering HotpotQA, with GPT-3.5 and GPT-4, and comparison results show that our framework outperforms the state-of-the-art training-free baselines. 

\end{abstract}

\section{Introduction}

Large Language Models (LLMs) have demonstrated strong reasoning capabilities across various tasks. More recently, researchers have applied LLMs in interactive tasks, where the LLMs iteratively make decisions based on new observations. These tasks range from performing online shopping and household chores to playing video games~\cite{reasoningsurvey,webshop}, which investigate the capabilities of LLMs as agents within complex environments.

In interactive tasks, ReAct~\cite{react} proposed an execution trajectory that interleaves reasoning, actions, and observations, which has been widely adopted by subsequent work~\cite{trainq,stateact}. However, as tasks become more complex, the interwoven reasoning and acting increases both the length and complexity of the context, which may lead to a decline in the model's performance~\cite{compositionality,distracted,longcontext}.
    
To mitigate this problem, researchers divide a complex task into subtasks with a planner, for example, statically planning the subtasks before execution~\cite{dprompt,pearl}, or dynamically determining the next subtasks during execution~\cite{adapt,adaplanner,tdag}. Although these subtasks can help reduce the complexity and length of context~\cite{thread,dprompt}, there are two potential issues: For one thing, subtasks might fail due to the lack of certain previous information, especially in tasks like multi-hop question answering~\cite{hotpotqa,beerqa}. For another, different subtasks may explore the same environment redundantly or look up information repeatedly, which increases the computational resource overhead. We argue that these are caused by certain subtasks, which are contextually related but executed within independent contexts. This relevance is unavoidable in many cases, because a complex task is difficult to be decomposed into largely unrelated subtasks. To address this issue, we propose a novel mechanism for exchanging information between subtasks, which we call \textit{interaction}.

In our \textit{interaction}, a subtask can send a request to a completed subtask, namely target subtask, to query specific information or trigger certain actions, accomplished by an in-context learning approach. Our \textit{interaction} faces several challenges. First, completed subtasks should support resumption from their execution trajectories to handle received interaction requests. Therefore, we introduced a memory module to store the trajectories of completed subtasks, so the target subtask is able to generate a response based on the retrieved trajectory. Second, to support flexible interaction initiation during subtask execution, we add a new action \textit{interact} to the action space of subtasks, and utilize a ReAct-style prompt to guide execution. Thus, the interaction will be triggered as other task-specific actions during execution.

\begin{figure}[t!]
\centering
\includegraphics[width=1\linewidth]{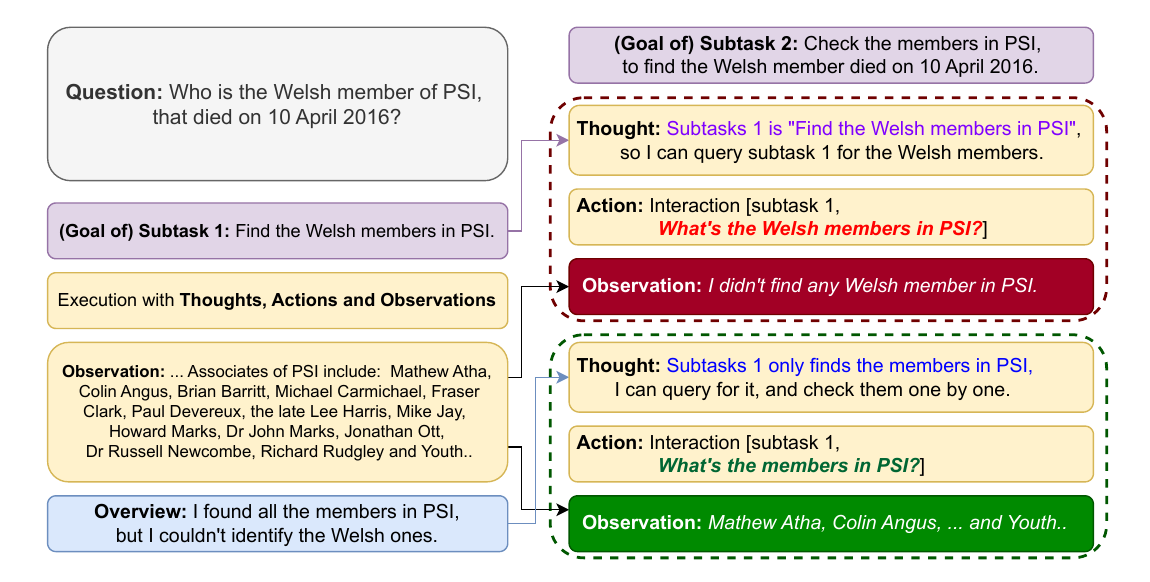}
\caption{Two interaction requests generated in question-answering task. \textbf{The red dashed box:} Request generated based on goal of subtask 1, which leads to an invalid request because subtasks 1 is not accomplished as expected. \textbf{The green dashed box:} Request generated based on overview of subtask 1, which is responded with effective information.}  
\label{fig:goal_or_overview}  
\end{figure}

Moreover, in our \textit{interaction}, a subtask determines the interaction target and request based on information from all completed subtasks. Describing this information is a challenge, as it must strike a balance bet`en conciseness, to minimize context overhead, and completeness, to provide essential information. Typically, the goal generated during planning is used to identify a subtask~\cite{rahl,adapt,tdag}, which we argue is merely a prior plan without actual outcomes, and may result in generating inappropriate interaction targets and requests. As demonstrated in the red dashed box in Figure~\ref{fig:goal_or_overview}, subtask 2 generates a request for information unknown to subtask 1 based on its goal. To resolve this, we add a new action \textit{finish~[overview]} to the action space of subtasks, which reports a short and precise description of its execution, called \textit{overview}. The overview encapsulates the execution process and result in a task-specific manner, guiding subsequent subtasks on determining interaction targets and requests. As demonstrated in the green dashed box in Figure~\ref{fig:goal_or_overview}, the overview describes the partial completion of subtask 1, making subtask 2 to successfully query the effective information.

To summarize, we introduce an LLM-based framework that enables interaction between subtasks within task decomposition settings. Our contributions are as follows:
\begin{itemize}

\item To handle the contextual relevance of subtasks within a task decomposition strategy, we propose a training-free framework with Interaction mechanism, named \textbf{I}nteractions \textbf{F}or task \textbf{D}ecomposition (IFD), which enables a subtask to perform interactions with completed subtasks in the form of sending requests.

\item To handle an interaction request, we propose a memory module to store execution trajectories of completed subtasks, from which the target subtask retrieves its trajectory to resume and respond to the request.

\item To help the executing subtask determine its interaction target and content, we present a new action to generate a concise and precise description of a subtask, called overview, which describes the execution process and outcomes of a completed subtask.

\end{itemize}

We evaluate IFD on interactive decision-making task WebShop~\cite{webshop} and data-grounded multi-hop question answering task HotpotQA~\cite{hotpotqa}. Compared with previous methods, IFD achieves leading performance, and outperforms the state-of-the-art training-free baselines. Moreover, we assess the contributions of interaction and overview through statistical analysis and ablation studies. The experiments demonstrate that interaction can lead to performance improvements across various tasks at a small cost.

\section{Related Work}

\subsection{Plan and Execute Strategy}

The success of Chain-of-Thought (CoT)~\cite{cot} demonstrates that LLMs can achieve more powerful reasoning capabilities by decomposing tasks. As tasks become increasingly complex, researchers introduced an independent planner to decompose tasks, thereby alleviating interference from complex execution trajectories~\cite{dprompt,swiftsage,pasprompt}. However, these approaches statically plan the execution process and struggle to handle situations where subtasks cannot be completed. To address this issue, researchers have proposed various dynamic planning methods. For instance, ADAPT~\cite{adapt} introduces a planner which further decomposes subtasks upon a subtask failed; AdaPlanner~\cite{adaplanner} adaptively adjusts its plan based on environmental feedback with LLM-based planner and refiner; TDAG~\cite{tdag} builds a multi-agent framework, where the planner dynamically decomposes or updates subtasks based on the result of completed subtask.

Existing planning and solving methods rarely consider information exchange between subtasks with independent contexts. THREAD~\cite{thread} treats model generation as a thread of execution, where child threads report results to parent threads upon completion. In some static planning methods~\cite{dprompt,compiler}, subtasks only pass execution results to the planner, which are used to replace the corresponding placeholders in subsequent subtasks. In contrast, IFD allows subtasks to interact with previous subtasks during execution, which establishes the information exchange between contextually related subtasks without additional training.

\subsection{LLM for Interactive Tasks}

Interactive tasks involve agents executing actions and processing observations. Researchers are exploring LLM-based framework in these tasks. Several methods~\cite{saycan,rvp} employ models to simulate or predict execution results, in response to potential environmental changes. However, these methods rely solely on the world knowledge of LLMs to simulate the environment, without incorporating actual environmental feedback. To address this issue, ReAct~\cite{react} integrates reasoning and execution in a single trajectories, enabling LLMs to update their plans based on environmental feedback. Subsequent work largely builds on ReAct’s approach and introduces optimizations~\cite{reflexion,laser,trainq,actre}. 

Inspired by the extensive adoption of ReAct~\cite{react} in interactive tasks, we consider using ReAct not only for interactions with the environment but also for interactions with prior subtasks. With a ReAct-style executor, interaction between subtasks is as seamless as with environments.

\section{Methodology}

\begin{figure*}[ht!]
\centering
\includegraphics[width=0.9\linewidth]{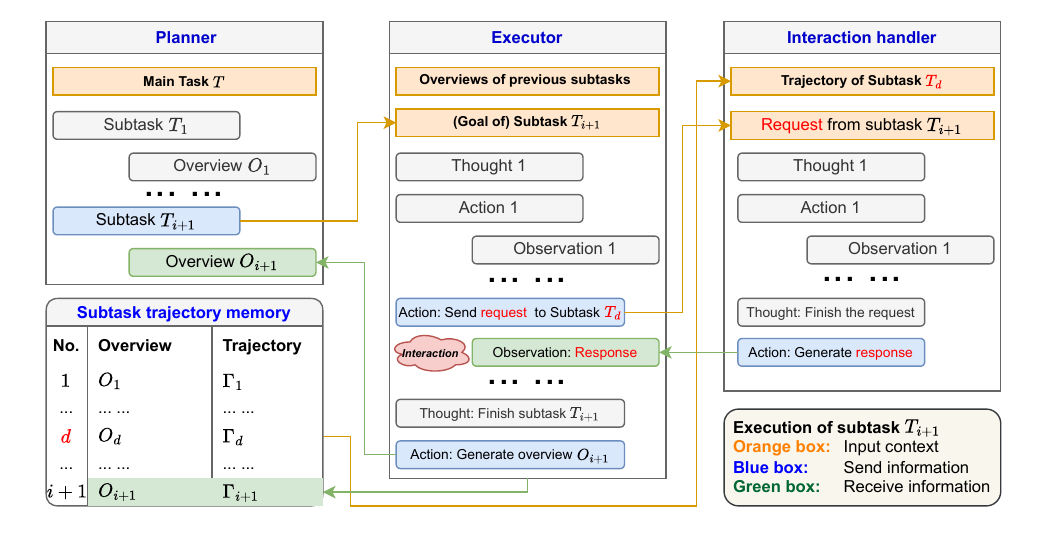}
\caption{The framework of IFD, illustrating the execution process of subtask $T_{i+1}$. The planner generates subtask $T_{i+1}$ based on the completed subtasks and their overviews. The executor processes the subtask $T_{i+1}$ in a ReAcT-style trajectory until generating an interaction with the subtask $T_d$ in the form of request. The interaction handler resumes subtask $T_d$ by loading its trajectory from the subtask trajectory memory, and produces the response with iterative execution. After receiving the response, the executor continues executing the subtask $T_{i+1}$ and generates an overview upon completion. Lastly, the trajectory and overview of the subtask \( T_{i+1} \) are stored in the subtask trajectory memory.}  
\label{fig:framework}  
\end{figure*}

We propose \textbf{I}nteractions \textbf{F}or task \textbf{D}ecomposition (IFD), a training-free framework with interaction between subtasks in form of request, as shown in Figure~\ref{fig:framework}. IFD consists of a dynamic planner, an executor, as well as a subtask trajectory memory and an interaction handler, both of which are used for interactions between subtasks.

\subsection{Planner}

IFD employs an LLM as planner to dynamically decompose a complex main task $T$ into easier subtasks $T_1, T_2, ..., T_n$. To dynamically generate a new subtask $T_{i+1}$, the planner considers the goals of previous subtasks $[T_i]$ as well as their overviews $[O_i]$, which are short descriptions of the execution process and results of the subtasks. 

\subsubsection{Dynamically planning with overviews}
Given a main task $T$, the planner dynamically generates $T_{i+1}$ as shown in Equation 1 and 2.
\begin{equation}
    T_1 = LLM_p(\mathcal{T}(T)),
\end{equation}
\begin{equation}
    T_{i+1} = LLM_p(\mathcal{T}(T, [T_i], [O_i])),\quad i \geq 1,
\end{equation}
where the $LLM_p(\cdot)$ is the LLM-based planner, the sequence $[T_i] = T_1, T_2, ..., T_i$ denotes the previous subtasks, and $[O_i] = O_1, O_2, ..., O_i$ corresponds to their overviews. We use a textual pattern $\mathcal{T}$ to structure $[T_i]$ and $[O_i]$ into a coherent description of the solution process of the main task $T$.

\subsection{Executor}

For a new subtask $T_{i+1}$, the executor iteratively generates thoughts and actions to accomplish it, and reports an overview $O_{i+1}$ upon completion. The action space of executor in IFD includes two additional actions, to initiate an interaction with a completed subtask, or generate an overview.

In particular, the execution trajectory $\Gamma^j_{i+1}$ of subtask $T_{i+1}$ with $j$ steps of LLM-based executor $LLM_e(\cdot)$ is described in Equation 3.
\begin{equation}
    \Gamma^j_{i+1} = \{t_1, a_1, o_1, t_2, a_2, o_2, ..., t_j, a_j, o_j\},
\end{equation}
where $t_j$ and $a_j$ represent thought and action generated by executor, and $o_j$ is the subsequent observation from the environment. 

\subsubsection{Interaction initiation by request}

To initiate an interaction with a previous subtask during execution, $interact$ is added to the action space $\mathcal{A}$ of executor, as described in Equation 4.
\begin{equation}
\mathcal{A} \leftarrow \mathcal{A} \cup interact(d, r),
\end{equation}
where $d$ indicates that the target subtask of this interaction, and $r$ represents a request in natural language.

When execution of a subtask $T_{i+1}$ begins, the executor is provided with overviews $[O_i]$ of previous subtasks, to formulate an appropriate target $d$ and request $r$, as shown in Equation 5.
\begin{equation}
    interact(d, r) = a_{j+1} = LLM_e([O_i], \Gamma^j \cup t_{j+1}),
\end{equation}

\subsubsection{Overview generation after execution}

When the subtask $T_{i+1}$ completed, the executor generates its overview $O_{i+1}$ as a short description of the execution process and result, which is accomplished by $finish$, another added action, as shown in Equation 6 and 7.
\begin{equation}
    \mathcal{A} \leftarrow \mathcal{A} \cup finish(O_{i+1}),
\end{equation}
\begin{equation}
    finish(O_{i+1}) = a_{j+1} = LLM_e([O_i], \Gamma^j \cup t_{j+1}),
\end{equation}
Besides task completed, the executor is required to generated overview when fail or step limit is reached, to report the encountered difficulties and acquired information.

\subsection{Interaction}

When an interaction is initiated by the executor, the target subtask resumes and generates a response. To achieve this, we introduce a subtask trajectory memory storing the trajectories of completed subtasks, and an LLM-based interaction handler for generating responses.

\subsubsection{Subtask trajectory memory} 

Subtask trajectory memory $\mathcal{M}$ is a dictionary that stores the execution trajectory $\Gamma_i$ and overview $O_i$ of each completed subtask $T_i$, labelled with their numbers $i$, as shown in Equation 8. 
\begin{equation}
    \mathcal{M} = \{(1, \Gamma_1, O_1), (2, \Gamma_2, O_2), ..., (i, \Gamma_i, O_i), \},
\end{equation} 
After a subtask $T_{i+1}$ finishes, its entire trajectory $\Gamma_{i+1}$ and generated overview $O_{i+1}$ will be added in memory $\mathcal{M}$.

\subsubsection{Interaction handler}

Consider an interaction with request $r$ between executing subtask $T_{i+1}$ and target subtask $T_d$. The LLM-based interaction handler $LLM_i(\cdot)$ resumes the subtask $T_d$ by loading its trajectory $\Gamma_d$ retrieved from memory $\mathcal{M}$ as context, and generates a response $res$ through an extra iterative execution, as shown in Equation 9 and 10.

\begin{equation}
    \Gamma_d' = LLM_i(r, \Gamma_d),
\end{equation}
\begin{equation}        
    finish(res) = a_j = last(\Gamma_d'),
\end{equation}

where $\Gamma_d'$ is the trajectory of the additional iterative execution process, which ends with an action $a_j$ that reports the response $res$ with $finish$. 

For simple requests that only involve querying for historical information from the trajectory $\Gamma_d$ of the subtask $d$, the trajectory $\Gamma_d'$ is a single-step reasoning containing a thought $t_1$ and an action $a_1$. For more complex requests that involve interaction with the external environment, the trajectory $\Gamma_d'$ is a sequence containing several thoughts, actions, and observations from the environment.

Finally, the response $res$ will be return to executing subtask $T_{i+1}$, as an observation $o_k$  following the action $a_k = interact(d, r)$.

\section{Experiments}

To evaluate \textbf{I}nteractions \textbf{F}or task \textbf{D}ecomposition (IFD), we compare it with baseline methods on an interactive web shopping environment WebShop~\cite{webshop}, and a Wikipedia-based multi-hop question answering benchmark HotpotQA~\cite{hotpotqa}. We conduct additional analysis and experiments on these two benchmarks to evaluate the impact of \textit{interaction} and \textit{overview}.

\begin{table}[ht!]
    \centering
    \begingroup
    \setlength{\tabcolsep}{3pt}
    \begin{tabular}{l>{\raggedleft\arraybackslash}p{1.4cm}>{\raggedleft\arraybackslash}p{1.7cm}}
        \toprule
        Method  & Success Rate (\%)& Average Reward (\%)\\
        \midrule
        ReAct~\cite{react}  & 29          & 42.1        \\
        ADAPT~\cite{adapt}  & 44          & 60.0        \\
        TDAG~\cite{tdag}    & 45          & 64.5        \\
        LATS~\cite{lats}    & 38          & 75.9        \\
        IFD (Ours)          &\textbf{48}  &\textbf{76.4}\\
        \bottomrule
    \end{tabular}
    \endgroup
    \caption{Success rate and average reward on webshop with baseline methods, with GPT-3.5-turbo as backbone model. }
    \label{tab:result_webshop}
\end{table}

\subsection{WebShop}

WebShop~\cite{webshop} is a simulated shopping website environment with 1.18 million real-world products, requiring an agent to purchase a product corresponding to a given instruction. The agent is challenged to interact with various webpages ranging from search result page to product description page, make decision based on natural language information, select available options and purchase the product that best matches the given instruction. Example 1 shows an instruction in WebShop.

\begin{example}
    I would like a 3-ounce bottle of bright citrus deodorant for sensitive skin, and a price lower than \$50.00.
\end{example}

\subsubsection{Baseline methods}

We compare IFD with four baseline methods: ReAct~\cite{react}, a classic prompt-based method, which our executor refers to; ADAPT~\cite{adapt} and TDAG~\cite{tdag}, two methods that dynamically decompose tasks based on the result of finish task; and LATS~\cite{lats}, an LLM framework based on Monte Carlo Tree Search, achieving state-of-the-art performance in terms of average reward.

\subsubsection{Setup}
Consistent with the settings of other baseline methods \cite{react,adapt,tdag}, we use the "all" version of the webshop benchmark for testing. The search returns 50 products, with 3 products displayed per page, and the agent is allowed to check up to 3 products.

Following ReAct~\cite{react}, we provide the agent with two types of actions to interact with the environment: (1) \textbf{search} [product], which triggers product search and returns the first page of results; (2) \textbf{click} [button], which clicks a button on the webpage to navigate to a new page, select product options, or purchase the product. To enable the \textit{interaction} between subtasks, we introduce two additional actions in IFD: (3) \textbf{interact} [target, request], which sends a request to a target subtask, and receives the result as observation from the target subtask; and (4) \textbf{finish} [overview], which finishes current subtask and returns \textit{overview} to the planner. 

We use two metrics for evaluation: average reward and success rate. The reward is provided by the environment, based on how well the purchased product matches the instruction, and is a value between 0 and 1. The success rate is the proportion of samples with a reward of 1.0 out of all test samples.

\subsubsection{Results}

Following the baseline methods~\cite{adapt,tdag}, we evaluate IFD on a test set of 100 examples. We calculate the success rate and average reward, and list them in Table~\ref{tab:result_webshop}. For the results of the baseline methods, we take the results of ReAct from the paper of TDAG~\cite{tdag}, and results of other baseline methods are directly taken from the original paper.

As shown in Table~\ref{tab:result_webshop}, IFD outperforms ReAct by 19\% in success rate and 34.3\% in average reward with a ReAct-Style executor. For ADAPT and TDAG, two baseline methods with dynamical planning, IFD outperforms them by 3\% in success rate and 11.9\% in average reward. Compared with LATS, IFD greatly exceeds it by 10\% in success rate, and achieves improvements in average reward.

In this experiment, IFD achieves state-of-the-art performance, which we argue is driven by \textit{interactions}. We analyse the 100 examples, identifying a total of 236 \textit{interactions}. To gain deeper insights into the impact of \textit{interaction}, we conducted statistical analyses and ablation experiments.

\begin{figure}[t!]
\centering
\includegraphics[width=\linewidth]{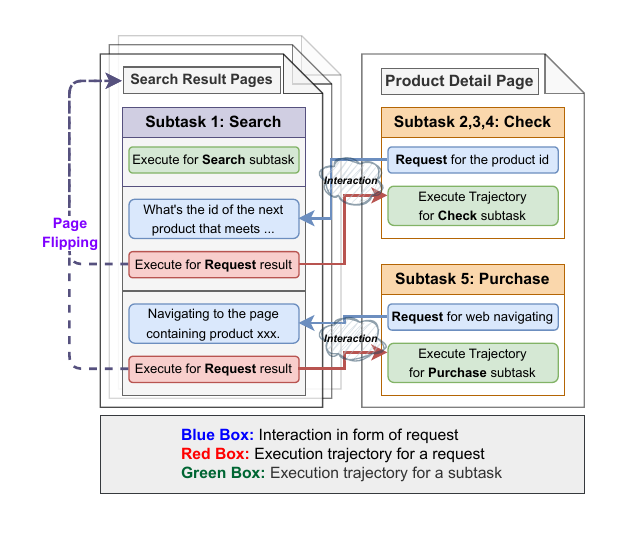}
\caption{Execution process of IFD in webshop, includes two kinds of \textit{interactions}. \textbf{The search subtask} searches for results. \textbf{The check subtasks} request the search subtask for the ID of the next product that meets specific requirements. \textbf{The buy subtask} requests the search subtask to navigate to the page containing the certain product. To respond to these request, the search subtask may need to perform page flipping.}  
\label{fig:webshop_execution}  
\end{figure}

\subsubsection{\textit{Interaction} increases reward}

In the execution process of IFD in WebShop, a subtask requires for the next products with specific requirement by interacting with the search subtask. The search subtask retrieves more results by flipping pages if no suitable one on the present page, as shown in Figure~\ref{fig:webshop_execution}. In contrast, all baseline methods with published prompts~\cite{react,adapt,lats} do not involve page flipping and only check up to three products on the first page, and buy the product with the highest predicted reward.

We examined the execution trajectories of IFD, and found that there are 8 examples where the purchased products are not listed on page 1. In these 6 examples, compared with purchasing the highest-rewarded product on the first page, the reward increases by an average of \textbf{0.21}, with \textbf{1} example's reward rising to 1.0, indicating success. 

To further verify the impact of \textit{interaction}, we conduct additional experiments on 200 examples, among which 11 involve purchasing products in pages after page 1. Compared with purchasing the highest-rewarded product on the first page, the reward increases by an average of \textbf{0.32}, with \textbf{3} examples' reward rising to 1.0, which indicates our \textit{interaction} indeed improves the results.

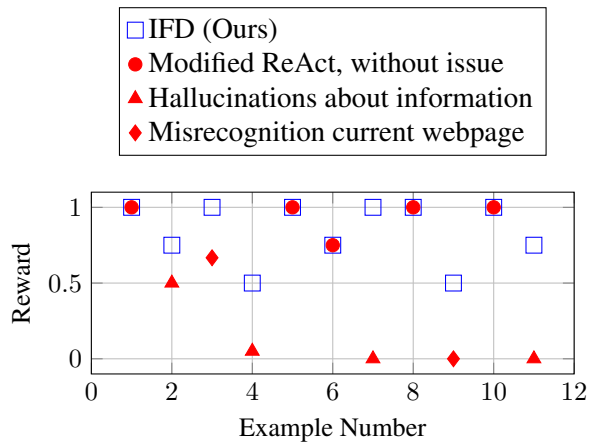
\begin{figure}[t!]
    \begin{tikzpicture}
      \begin{axis}[
        xlabel={Example Number},
        ylabel={Reward},
        legend style={at={(0.5,1.2)}, anchor=south, 
        font=\large},
        legend cell align={left},
        grid=both,
        width=8cm, height=4cm,
      ]
      
        \addplot[
          only marks,
          mark=square,
          color=blue,
          mark size=3pt
        ]
        coordinates {
          (1,1.0) (2,0.75) (3,1.0) (4,0.5) (5,1.0) (6,0.75) (7,1.0) (8,1.0)(9,0.5) (10,1.0) (11,0.75)
        };
        \addlegendentry{IFD (Ours)}
        
        \addplot[
          only marks,
          mark=*,
          color=red,
          mark size=2.5pt
        ]
        coordinates {
          (1,1.0) (5,1.0) (6,0.75) (8,1.0) (10,1.0)
        };
        \addlegendentry{Modified ReAct, without issue}
    
        \addplot[
          only marks,
          mark=triangle*,
          color=red,
          mark size=3pt
        ]
        coordinates{
          (2,0.5) (4,0.05) (7,0.0) (11,0.0)
        };
        \addlegendentry{Hallucinations about information}
    
        \addplot[
          only marks,
          mark=diamond*,
          color=red,
          mark size=3pt
        ]
        coordinates {
          (3,0.667) (9,0.0)
        };
        \addlegendentry{Misrecognition  current webpage}
      \end{axis}
    \end{tikzpicture}
    \caption{The React method with page flipping functionality encounters issues on several examples.}
    \label{fig:result_page_flipping}
\end{figure}

\subsubsection{\textit{Interaction} reduces context complexity}

In IFD, the introduction of \textit{interaction} enables each subtask to be executed on a single webpage, which reduces the complexity of the execution context. Figure~\ref{fig:webshop_execution} illustrates the \textit{interactions} in form of requests in the execution process of IFD in WebShop. The search subtask executes on the search results pages and processes subsequent received \textit{interactions}. The check and purchase subtasks perform operations which may involve page flipping through \textit{interactions}, and receive only the result as observation, with the additional context introduced by page flipping being invisible to them. 

Although other methods might also support page flipping operations by modifying prompts, they may struggle to handle the increased contextual complexity introduced by page transitions. To validate this point, we modified ReAct to support page flipping and compared it with IFD. We designed the prompt for ReAct based on IFD, ensuring it supports page flipping. We compared IFD with the modified ReAct method on the 11 examples mentioned above that buy a product in pages after page 1, using GPT-4o model.

Figure~\ref{fig:result_page_flipping} shows the result. The modified ReAct encounters issues in 55\% of the samples. Of these, 37\% generate hallucinations about the webpage information, such as incorrectly identifying product attributes, while 18\% involve misrecognizing the current webpage, such as attempting to perform a purchase action on a page without a 'purchase' button. Moreover, the modified ReAct does not outperform IFD on these 11 examples. The experimental results indicate that IFD can better handle frequent webpage switching.

\subsection{HotpotQA}

HotPotQA~\cite{hotpotqa} is an interactive Wikipedia-based dataset consisting of 113,000 question-and-answer pairs, where an agent is challenged to search for external documents, and reason across multiple supporting information. Example 2 shows a question in HotpotQA.

\begin{example}
    when was the former Indian cricketer who made his ODI debuts during the 2000 ICC KnockOut Trophy born?
\end{example}

\subsubsection{Baseline methods}

Since methods with dynamic task decomposition are rarely evaluated on multi-hop question answering, we adopt ReAct~\cite{react} and two recent LLM-based methods as baseline approaches, LEAP~\cite{leap} and the method presented in paper~\cite{trainq}, which we refer to as Q model. Additionally, we compare IFD with two approaches that were refined through multiple attempts~\cite{reflexion,retroformer}, which achieve state-of-the-art performance in LLM-based methods.

\subsubsection{Setup}

The previous work~\cite{react,reflexion,leap} commonly operate in a question-only setup, where the agents only receive the question as input without accessing the support paragraphs, and interact with a Wikipedia API or retriever for external facts. We follow the settings in ReAct~\cite{react} in our experiments.

In detail, the agent is provided with three types of actions to interact with the environment: (1) \textbf{search} [entity], which searches for the entity on Wikipedia and returns the first paragraph if it exists; otherwise, it returns similar entities. (2) \textbf{lookup} [keyword], which returns the next sentence containing the keyword from the page successfully found by \textbf{search}. (3) \textbf{answer} [answer], which submits the answer and finishes the task. To enable \textit{interaction} between subtasks, we introduce two additional actions in IFD: (4) \textbf{interact} [target, request], which sends a request, typically a query, to target subtask, and receives the result as observation, and (5) \textbf{finish} [overview], which finishes current subtask and returns its \textit{overview}. We adopt the success rate based on exact match as the evaluation metric.

\begin{table}[t]
    \centering
    \begin{tabular}{lrr}
        \toprule
        Sigle-pass Method      & GPT-4o-mini & GPT-4       \\
        \midrule
        ReAct~\cite{react}     & 31          & 44          \\
        LEAP~\cite{leap}       & -           & 40          \\
        Q model~\cite{trainq} *& \textbf{44} & 50          \\
        IFD (Ours)             & 43          & \textbf{52} \\
        \bottomrule
    \end{tabular}
    \caption{Success rate (\%) on HotpotQA, compared with methods with sigle-pass execution. The Methods marked with an asterisk (*) require additional training.}
    \label{tab:result_hotpotqa_single}

    \vspace{0.5cm}

    \centering
    \begin{tabular}{lrr}
        \toprule
        Method with Attempts            & Retries & GPT-4       \\
        \midrule
        Reflexion~\cite{reflexion}      & 1       & 46          \\
                                        & 4       & 52          \\
                                    \hline
        Retroformer~\cite{retroformer} *& 1       & 51          \\
                                        & 4       & \textbf{54} \\
                                    \hline
        IFD (Ours)                      & 0       & 52          \\
        \bottomrule
    \end{tabular}
    \caption{Success rate (\%) on HotpotQA, compared with methods with multiple attempts. The Methods marked with an asterisk (*) require additional training.}
    \label{tab:result_hotpotqa_attempts}
\end{table}

\subsubsection{Results}

Following the baseline methods~\cite{leap,trainq,reflexion,retroformer}, we evaluate IFD on a test set of 100 examples. We calculate the success rate, and list the results compared with sigle-pass methods in Table~\ref{tab:result_hotpotqa_single}, the results compared with multiple attempts in Table~\ref{tab:result_hotpotqa_attempts}. For the results of the baseline methods, we take the results of ReAct from the paper of Q model~\cite{trainq} because the original paper does not use gpt-4 series model, and results of Reflexion from the paper of Retroformer~\cite{retroformer} because the original paper does not report the number of attempts. Except for these two, we report the results from their original paper.

As shown in Table~\ref{tab:result_hotpotqa_single}, compared with single-pass methods, IFD outperforms ReAct by 12\% on GPT-4o-mini. On GPT-4, IFD exceeds ReAct by 8\%, and outperforms LEAP by 12\%, which is the only method based on COT rather than ReAct. Compared with the Q model which requires additional training, IFD outperformed it by 2\% on GPT-4, while being only 1\% inferior to it on GPT-4o-mini.

As shown in Table~\ref{tab:result_hotpotqa_attempts}, compared with methods with retry attempts, IFD outperforms Reflexion with 1 retry by 6\%, and on par with Reflexion using 4 retries. IFD surpasses Retroformer with 1 retry by 1\%, and falls short by 2\% compared with Retroformer with 4 retries.

In this experiment, IFD demonstrates competitive performance, achieving state-of-the-art results among training-free methods. We attribute this success to the introduction of \textit{interaction}, as analysed below.

\subsubsection{\textit{Interaction} enables multi-hop reasoning with subtasks}

We analyse 100 execution trajectories of IFD on GPT-4o-mini. On average, each example involves retrieving and processing approximately 3.08 Wikipedia paragraphs per trajectory in the absence of task decomposition. Moreover, IFD divides each example into an average of 3.1 subtasks, reducing the number of paragraphs retrieved per subtask to just $3.08/3.1 = 0.99$. IFD reduces contextual complexity through task decomposition, enabled by the introduction of \textit{interaction} for information exchange between subtasks.

In HotpotQA, a question requires multiple reasoning steps, with each step potentially relying on information from previous ones. While task decomposition lends itself to transforming multi-hop question answering into sequential single-step reasoning tasks, existing methods are rarely evaluated on multi-hop reasoning benchmark. We argue that the inability of subtasks to access information from previous subtasks leads to an unsuitability to multi-hop reasoning. In contrast, IFD addresses this issue through \textit{interaction}: \textbf{interact} [target subtask, Tell me about ...], enabling the distribution of execution trajectories across subtasks in multi-hop reasoning.

\subsubsection{\textit{Overview} promotes \textit{Interaction}}

\begin{table}
    \centering
    \begin{tabular}{lrr}
        \toprule
                                        & \textit{Overview} & Goal \\
        \midrule
        Total \textit{interactions}     & 89    & 180   \\
        Failed \textit{interactions}    & 3     & 26    \\
        Failed rate (\%)                & 3.4   & 14.4  \\
        \bottomrule
    \end{tabular}
    \caption{\textit{Interaction} failed rate under two setting: providing subtasks with the \textit{overviews} or goals of completed subtasks.}
    \label{tab:result_goal_and_performance}
\end{table}

In IFD, the current subtask determines the \textit{interaction}'s target and request based on the \textit{overview} of completed subtasks. To investigate the role of \textit{overview} in \textit{interaction}, we conducted an ablation experiment.

We replace the \textit{overviews} of the completed subtasks with their goals, and conduct an experiment on 100 examples, using GPT-4o-mini as the backbone model. We count the number of all \textit{interactions}, including successful and failed \textit{interactions}. The failed \textit{interactions} refer to requesting for unknown information.

The results are shown in Table~\ref{tab:result_goal_and_performance}. The \textit{interaction} failed rate is 14.4\% when representing completed subtasks with their goals, compared with the failed rate 3.4\% of our \textit{overview}. This confirms that \textit{overview} can assist in generating effective \textit{interaction} target and request.

Notably, compared with providing goals of completed subtasks, the number of \textit{interactions} is greatly reduced by 50.6\% when using \textit{overview}. We attribute this reduction to two main reasons: First, the executor may adjust the \textit{interaction} target or query to retry after a failed \textit{interaction}. Second, some \textit{overviews} already include information needed for subsequent subtasks, thus eliminating the need for further \textit{interactions}. This result indicates that \textit{overviews} can reduce unnecessary \textit{interactions}, thereby saving computational resources.

\section{Discussion}
In the Discussion, we explore the versatility and adaptability of our interaction.

We apply interaction across different benchmarks with the same two additional actions: \textit{interact [target, request]} and \textit{finish [overview]}. In WebShop, interaction allows each subtask to be executed within a single webpage environment, with cross-page operations performed through requests, to reduce the complexity of execution trajectories. For example, the purchase subtask navigates to the page containing the target product item via sending a request to the search subtask, rather than performing page flipping operations. In HotpotQA, interaction transfers information between subtasks with independent contexts, enabling multi-hop reasoning based on information retrieved from previous subtasks. The same implementation of interaction resulting in varying effects, demonstrating its versatility. We believe that it can become a common component for solving complex tasks in the future.

The previous methods based on task decomposition~\cite{adaplanner,adapt,thread,tdag} are mainly evaluated on two interactive environments, ALFworld~\cite{alfworld} and WebShop~\cite{webshop}, with a few~\cite{dprompt,compiler} using question-answering benchmarks, where methods are challenged to access information from previous subtasks. Although our framework can adapt to various complex tasks by modifying few-shot prompts, it's not suitable for some tasks where interactions between subtasks is unnecessary~\cite{alfworld,intercode}. Therefore, our method is not evaluated on ALFWorld, where a task can be completed without any interaction between subtasks, as shown in Example 3.

\begin{example}
    Task: wash a cup and put it in the shelf.
\end{example}

\section{Conclusion}

Task decomposition is widely adopted for solving complex tasks using LLMs. However, when the decomposed subtasks are contextually related, task execution may fail due to the independence of their execution contexts. To address this issue, we propose \textbf{I}nteractions \textbf{F}or task \textbf{D}ecomposition (IFD), a training-free framework that incorporates an interaction mechanism to enable information transfer between related subtasks. We evaluate IFD in interactive task and multi-hop reasoning, and verify roles of the two key components: \textit{interaction} and \textit{overview}. Experimental results demonstrate that IFD achieves leading performance on these benchmarks, and generalizes well in both decision-making and reasoning tasks. Our future work may involve integrating interaction mechanisms into LLMs through fine-tuning, investigating better structured representations for the overview component, and applying interaction to a greater variety of tasks.

\newpage

\bibliographystyle{named}
\bibliography{ijcai25}

\end{document}